\documentclass[10pt,twocolumn,letterpaper]{article}

\usepackage{wacv}
\usepackage{times}
\usepackage{epsfig}
\usepackage{graphicx}
\usepackage{amsmath}
\usepackage{amssymb}

\usepackage{color}
\usepackage{dsfont}
\usepackage{algorithm}
\usepackage{algorithmicx}
\usepackage{bm}
\usepackage{diagbox}

\usepackage{relsize}

\usepackage{adjustbox}
\usepackage{algpseudocode}
\usepackage{multirow}
\usepackage{array}
\usepackage{booktabs} 

\DeclareMathOperator*{\argmax}{arg\,max}

\usepackage[pagebackref=true,breaklinks=true,letterpaper=true,colorlinks,bookmarks=false]{hyperref}

\wacvfinalcopy 

\def\wacvPaperID{184} 

\ifwacvfinal\fi
\begin{document}

\title{Semi-Supervised 3D Abdominal Multi-Organ Segmentation via Deep Multi-Planar Co-Training}

\author{Yuyin Zhou$^1$~~Yan Wang$^1$~~Peng Tang$^2$~~Song Bai$^3$~~Wei Shen$^1$~~Elliot K. Fishman$^4$~~Alan Yuille$^1$ \\
$^1$The Johns Hopkins University~~~$^2$Huazhong University of Science and Technology\\
$^3$University of Oxford~~~~$^4$The Johns Hopkins University School of Medicine\\
{\tt\small\{zhouyuyiner,wyanny.9,tangpeng723,songbai.site,shenwei1231\}@gmail.com}\\
{\tt\small alan.l.yuille@gmail.com, efishman@jhmi.edu}
}

\maketitle
\ifwacvfinal\thispagestyle{empty}\fi

\begin{abstract}
In multi-organ segmentation of abdominal CT scans, most existing fully supervised deep learning algorithms require lots of voxel-wise annotations, which are usually difficult, expensive, and slow to obtain.
In comparison, massive unlabeled 3D CT volumes are usually easily accessible.
Current mainstream works to address semi-supervised biomedical image segmentation problem are mostly graph-based. By contrast, deep network based semi-supervised learning methods have not drawn much attention in this field. In this work, we propose Deep Multi-Planar Co-Training (DMPCT), whose contributions can be divided into two folds: 
1) The deep model is learned in a co-training style which can mine consensus information from multiple planes like the sagittal, coronal, and axial planes;
2) Multi-planar fusion is applied to generate more reliable pseudo-labels, which alleviates the errors occurring in the pseudo-labels and thus can help to train better segmentation networks. Experiments are done on our newly collected large dataset with $100$ unlabeled cases as well as $210$ labeled cases where $16$ anatomical structures are manually annotated by four radiologists and confirmed by a senior expert. The results suggest that DMPCT significantly outperforms the fully supervised method by more than $4\%$ especially when only a small set of annotations is used.
\end{abstract}

\section{Introduction}
\label{Introduction}

Multi-organ segmentation of radiology images is a critical task which is essential to many clinical applications such as computer-aided
diagnosis, computer-aided surgery, and radiation therapy.
Compared with other internal human structures like brain or heart,
segmenting abdominal organs appears to be much more challenging due to the low contrast and high variability of shape in CT images.
In this paper, we focus on the problem of multi-organ segmentation in abdominal regions,~\emph{e.g.}, liver, pancreas, kidney, \emph{etc.}

Fully supervised approaches can usually achieve high accuracy with a large labeled training set which consists of pairs of radiology images as well as their corresponding pixel-wise label maps. However, it is quite time-consuming and costly to obtain such a large training set especially in the medical imaging domain due to the following reasons: 1) precise annotations of radiology images must be hand annotated by experienced radiologists and carefully checked by additional experts and 2) contouring organs or tissues in 3D volumes requires tedious manual input. By contrast, large unannotated datasets of CT images are much easier to obtain. Thereby our study mainly focuses on multi-organ segmentation in a semi-supervised fashion, \emph{i.e.}, how to fully leverage unlabeled data to boost performance, so as to alleviate the need for such a large annotated training set.


In the biomedical imaging domain, traditional methods for semi-supervised learning usually adopt graph-based methods \cite{ciurte2014semi,Gu2017MICCAI} with a clustering assumption to segment pixels (voxels) into meaningful regions, \emph{e.g.}, superpixels. These methods were studied for tissue or anatomical structures segmentation in 3D brain MR images, ultrasound images, \emph{etc}. Other machine learning methods such as kernel-based large margin algorithms \cite{Qin2016MICCAI} have been suggested for white matter hyperintensities segmentation. Although widely applied to biomedical imaging segmentation tasks in the past decade, the traditional methods cannot always produce a satisfactory result due to the lack of advanced techniques. 

With the recent advance of deep learning and its applications \cite{Ref:Krizhevsky2012, tang2017multiple, tang2018pcl, tang2018weakly}, fully convolutional networks (FCNs) \cite{Long_2015_Fully} have been successfully applied to many biomedical segmentation tasks such as neuronal structures segmentation \cite{chen2016deep,Ciresan_2012_Deep,Ronneberger_2015_UNet,shen2017multi}, single organ segmentation \cite{Roth_2016_Spatial,zhou2017fixed, zhou2017deep}, and multi-organ segmentation \cite{roth2017hierarchical,wang2018training} in a fully supervised manner. Their impressive performances have shown that we are now equipped with much more powerful techniques than traditional methods. Nevertheless, network-based semi-supervised learning for biomedical image segmentation has not drawn enough attention. The current usage of deep learning for semi-supervised multi-organ segmentation in the biomedical imaging domain is to train an FCN on both labeled and unlabeled data, and alternately update automated segmentations (pseudo-labels) for unlabeled data and the network parameters \cite{bai2017semi}. However, if an error occurs in the initial pseudo-label of the unlabeled data, the error will be reinforced by the network during the following iterations. How to improve the quality of pseudo-labels for unlabeled data hence becomes a promising direction to alleviate this negative effect.

In this paper, we exploit the fact that CT scans are high-resolution three-dimensional volumes which can be represented by multiple planes, \emph{i.e.}, the axial, coronal, and sagittal planes. 
Taking advantages of this multi-view property, we propose Deep Multi-Planar Co-Training (DMPCT), a systematic EM-like semi-supervised learning framework. DMPCT consists of a teacher model, a multi-planar fusion module, and a student model. While the teacher model is trained from multiple planes separately in a slice-by-slice manner with a few annotations, the key advantage of DMPCT is that it enjoys the additional benefit of continuously generating more reliable pseudo-labels by the multi-planar fusion module, which can afterward help train the student model by making full usage of massive unlabeled data. 
As there are multiple segmentation networks corresponding to different planes in the teacher model and the student model, co-training \cite{blum1998combining, qiao2018deep} is introduced so that these networks can be trained simultaneously in our unified framework and benefit from each other.
We evaluate our algorithm on our newly collected large dataset and observe a significant improvement of $4.23\%$ compared with the fully supervised method. At last, as DMPCT is a generic and flexible framework, it can be envisioned that better backbone models and fusion strategies can be easily plugged into our framework. Our unified system can be also practically useful for current clinical environments due to the efficiency in leveraging massive unlabeled data to boost segmentation performance. 

\section{Related Work}

\noindent\textbf{Fully-supervised multi-organ segmentation.} Early studies of abdominal organ segmentation focused on atlas-based methods \cite{Iglesias2015,Chu2013,Wolz2013}. The frameworks are usually problematic because 1) they are not able to capture the large inter-subject variations of abdominal regions and 2) computational time is tightly dependent on the number of atlases.
Recently, learning-based approaches with relatively large dataset have been introduced for multi-organ segmentation \cite{gibson2017towards,roth2018multi, brosch2018foveal}. Especially, deep Convolutional Neural Networks (CNNs) based methods have achieved a great success in the medical image segmentation \cite{roth2017hierarchical, chen2017towards,gibson2018automatic,wang2018abdominal, wang2018training, zhou2016three, hu2017automatic} in the last few years. Compared with multi-atlas-based approaches, CNNs based methods are generally more efficient and accurate. CNNs based methods for multi-organ segmentation can be divided into two major categories: 3D CNNs \cite{roth2017hierarchical, chen2017towards,gibson2018automatic} based and 2D CNNs \cite{wang2018abdominal, wang2018training, zhou2016three, hu2017automatic} based. 3D CNNs usually adopt the sliding-window strategy to avoid the \emph{out of memory} problem, leading to high time complexity. Compared with 3D CNNs, 2D CNNs based algorithms can be directly end-to-end trained using 2D deep networks, which is less time-consuming.

\begin{figure*}[t]
\begin{center}
    \includegraphics[width=0.85\linewidth]{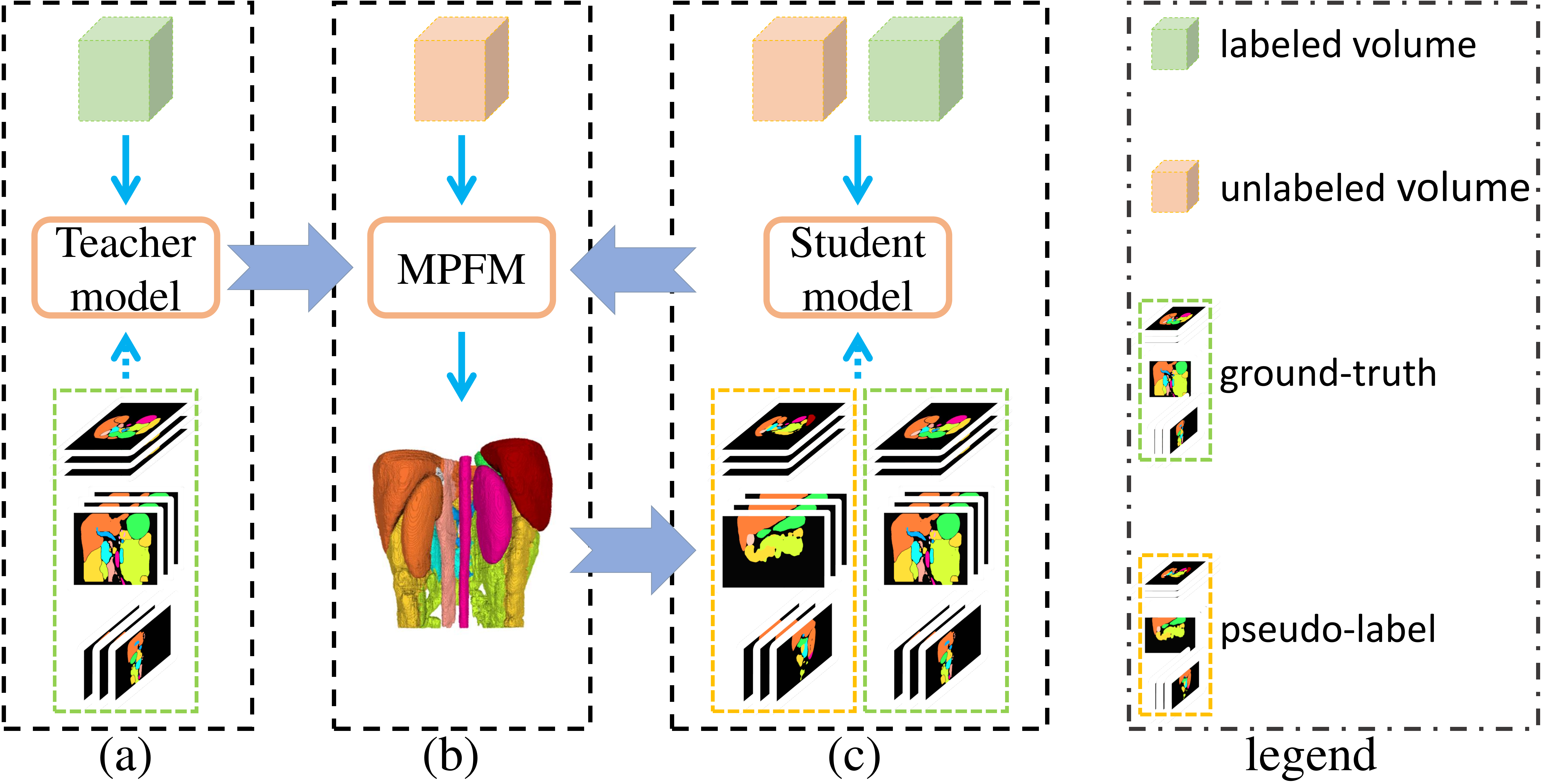}
\end{center}
\vspace{-1ex}
\caption{Illustration of the Deep Multi-Planar Co-Training (DMPCT) framework.
(a) We first train a teacher model on the labeled dataset.
(b) The trained model is the used to assign pseudo-labels to the unlabeled data using our multi-planar fusion module as demonstrated in Figure \ref{Fig:MVEM}.
(c) Finally, we train a student model over the union of both the labeled and the unlabeled data.
Step (b) and (c) are performed in an iterative manner.
}
\label{Fig:MVCT}
\end{figure*}

\vspace{0.1cm}
\noindent\textbf{Semi-supervised learning.}
The most commonly used techniques for semi-supervised learning include self-training~\cite{rosenberg2005semi,radosavovic2017data}, co-training~\cite{blum1998combining}, multi-view learning~\cite{xu2013survey} and graph-based methods~\cite{blum2001learning, wang2013semi}. 

In self-training, the classifier is iteratively re-trained using the training set augmented by adding the unlabeled data with their own predictions. The procedure repeated until some convergence criteria are satisfied. In such case, one can imagine that a classification mistake can reinforce itself. Self-training has achieved great performances in many computer vision problems~\cite{rosenberg2005semi,radosavovic2017data} and recently has been applied to deep learning based semi-supervised learning in the biomedical imaging domain~\cite{bai2017semi}. 

Co-training~\cite{blum1998combining} assumes that (1) features can be split into two independent sets and (2) each sub-feature set is sufficient to train a good classifier. During the learning process, each classifier is retrained with the additional training examples given by the other classifier. Co-training utilizes multiple sets of independent features which describe the same data, and therefore tends to yield more accurate and robust results than self-training~\cite{sousa2017comparison}. Multi-view learning~\cite{xu2013survey}, in general, defines learning paradigms that utilize the agreement among different learners. Co-training is one of the earliest schemes for multi-view learning. 

Graph-based semi-supervised methods define a graph where the nodes are labeled and unlabeled examples in the dataset, and edges reflect the similarity of examples. These methods have been widely adopted in non-deep-learning based semi-supervised learning algorithms in the biomedical imaging domain~\cite{ciurte2014semi,Gu2017MICCAI, Qin2016MICCAI}.

Different from other methods, our work tactfully embeds the multi-view property of 3D medical data into the co-training framework, which is simple and effective.

\section{Deep Multi-Planar Co-Training}
\label{Approach}

We propose Deep Multi-Planar Co-Training (DMPCT), a semi-supervised multi-organ segmentation method which exploits multi-planar information to generate pseudo-labels for unlabeled 3D CT volumes.
Assume that we are given a 3D CT volume dataset $\mathcal{S}$ containing $K$ organs. This includes labeled volumes $\mathcal{S}_{\text{L}}=\{(\mathbf{I}_m,\mathbf{Y}_m)\}_{m=1}^l$ and unlabeled volumes $\mathcal{S}_{\text{U}}=\{\mathbf{I}_m\}_{m=l+1}^M$, where $\mathbf{I}_m$ and $\mathbf{Y}_m$ denote a 3D input volume and its corresponding ground-truth segmentation mask. $l$ and $M-l$ are the numbers of labeled and unlabeled volumes, respectively. Typically $l \ll M$.
As shown in Figure~\ref{Fig:MVCT},
DMPCT involves the following steps:
\begin{itemize}
\item \textbf{Step 1:} train a \emph{teacher model} on the manually labeled data $\mathcal{S}_{\text{L}}$ in the fully supervised setting (see Sec. \ref{baseline}).
\item \textbf{Step 2:} the trained model is then used to assign pseudo-labels $\{\hat{\mathbf{Y}}_m\}_{m=l+1}^M$ to the unlabeled data $\mathcal{S}_{\text{U}}$ by fusing the estimations from all planes (see Sec. \ref{Test}).
\item \textbf{Step 3:} train a \emph{student model} on the union of the manually labeled data and automatically labeled data $\mathcal{S}_{\text{L}}\cup\{(\mathbf{I}_m,\hat{\mathbf{Y}}_m)\}_{m=l+1}^M$ (see Sec. \ref{Approach:Student}).
\item \textbf{Step 4:} perform step 2 \& 3 in an iterative manner.
\end{itemize}

\begin{figure*}[t]
\centering
\includegraphics[width=0.81\linewidth]{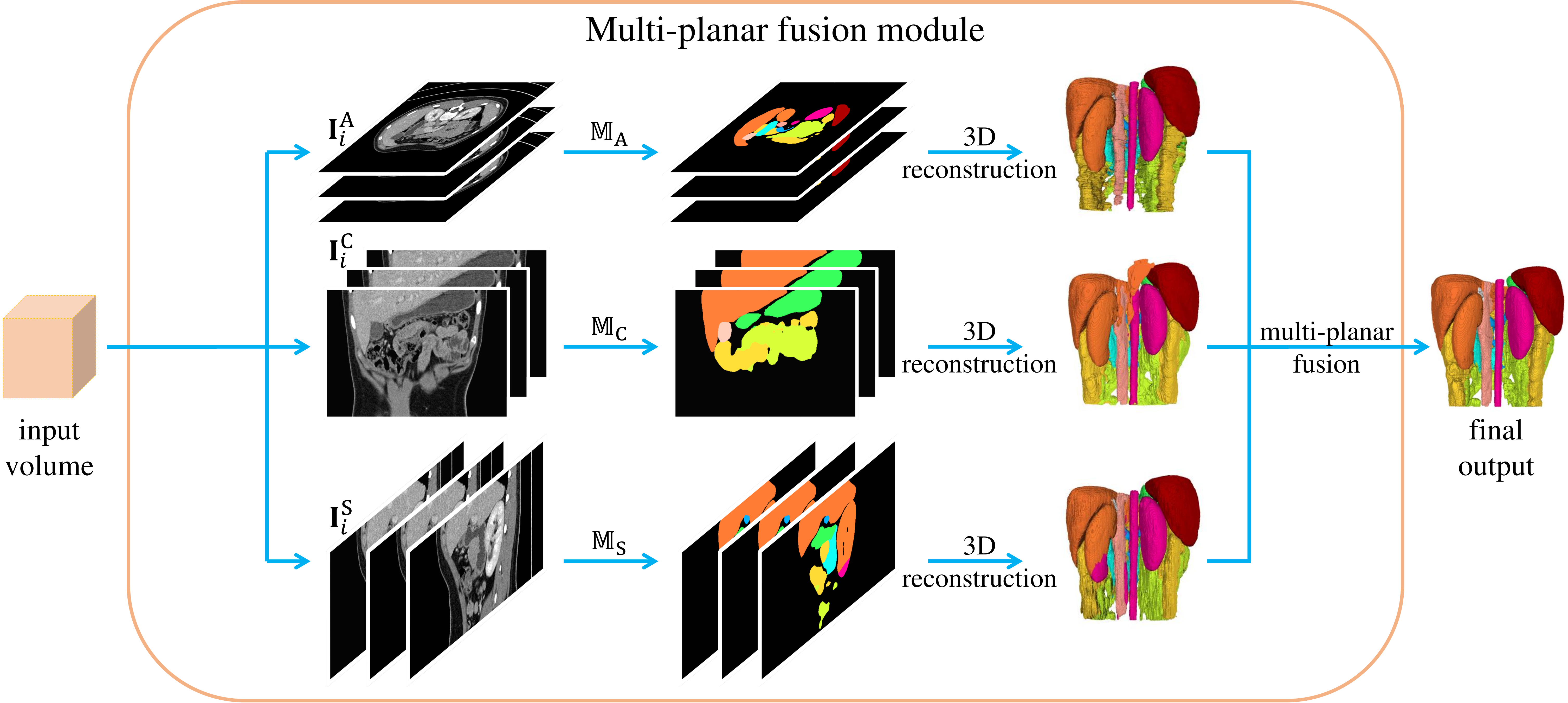}
\caption{Illustration of the multi-planar fusion module, where the input 3D volume is first parsed into 3 sets of slices along the sagittal, coronal, and axial planes to be evaluated respectively. Then the final 3D estimation is obtained by fusing predictions from each individual plane.}
\label{Fig:MVEM}
\end{figure*}

\subsection{Teacher Model}
\label{baseline}

We train the teacher model on the labeled dataset $\mathcal{S}_{\text{L}}$. By splitting each volume and its corresponding label mask from the sagittal (S), coronal (C), and axial (A) planes, we can get three sets of 2D slices, \emph{i.e.}, $\mathcal{S}_{\text{L}}^{V} = \{(\mathbf{I}_n^{V}, \mathbf{Y}_n^{V})\}_{n=1}^{N_{V}}$, ${V}\in\{\text{S},\text{C},\text{A}\}$, where $N_{V}$ is the number of 2D slices obtained from plane ${V}$.
We train a 2D-FCN model (we use ~\cite{Long_2015_Fully} as our reference CNN model throughout this paper) to perform segmentation from each plane individually. 

Without loss of generality, let $\mathbf{I}^V \in \mathbb{R}^{W \times H}$ and $\mathbf{Y}^V = \{y_i^V\}_{i=1}^{W \times H}$ denote a 2D slice and its corresponding label mask in $\mathcal{S}_{\text{L}}^{\text{V}}$, where $y_i^V\in\{0,1,\ldots,K\}$ is the organ label (0 means background) of the $i$-th pixel in $\mathbf{I}^V$. Consider a segmentation model $\mathbb{M}^V:{\hat{\mathbf{Y}}}={\mathbf{f}\!\left(\mathbf{I}^V; \mathcal{\theta}\right)}$, where $\mathcal{\theta}$ denotes the model parameters and $\hat{\mathbf{Y}}$ denotes the prediction for $\mathbf{I}^V$. Our objective function is
\begin{equation}
\mathcal{L}(\mathbf{I}^V, \mathbf{Y}^V; \mathcal{\theta}) = -\frac{1}{W \times H} \bigg[\sum\limits_{i=1}^{W \times H} \sum\limits_{k=0}^K \mathds{1}  (y_i^V=k) \log p_{i,k}^V \bigg],
\end{equation}
where $p_{i,k}^V$ denotes the probability of the $i$-th pixel been classified as label $k$ {on} 2D slice $\mathbf{I}^V$ and $\mathds{1} (\cdot) $ is the indicator function. We train the teacher model by optimizing  $\mathcal{L}$ w.r.t.~$\mathcal{\theta}$ by stochastic gradient descent.

\subsection{Multi-Planar Fusion Module}
\label{Test}
Given a well-trained teacher model $\{\mathbb{M}^V|V \in\{\text{S}, \text{C}, \text{A}\}\}$, our goal of the multi-planar fusion module is to generate the pseudo-labels $\{\hat{\mathbf{Y}}_m\}_{m=l+1}^M$ for the unlabeled data $\mathcal{S}_{\text{U}}$.
We first make predictions on the 2D slices from each plane and then reconstruct the 3D volume by stacking all slices back together. Several previous studies \cite{heckemann2006automatic,rohlfing2004evaluation,gift_cvpr,gift_tmm} suggest that combining predictions from multiple views can often improve the accuracy and the robustness of the final decision since complementary information can be exploited from multiple views simultaneously.
Thereby, the fused prediction from multiple planes is superior to any estimation of a single plane. The overall module is shown in Figure~\ref{Fig:MVEM}. 

\begin{figure*}[t]
\begin{center}
    \includegraphics[width=0.81\linewidth]{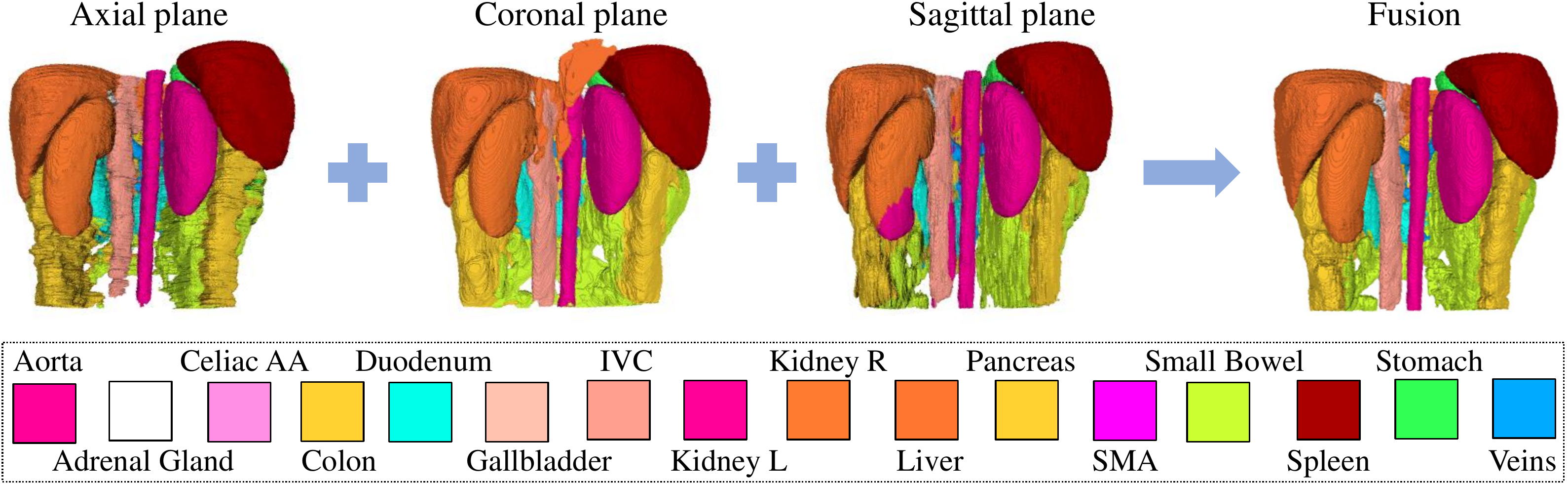}
\end{center}
\vspace{-1ex}
\caption{
    An example of 3D predictions reconstructed from the sagittal, coronal, and axial planes as well as their fusion output. Estimations from single planes are already reasonably well, whereas the single fusion outcome is superior to estimation from any single plane.}
\label{Fig:fusion}
\end{figure*}

More specifically, majority voting is applied to fuse the hard estimations by seeking an agreement among different planes. If the predictions from all planes do not agree on a voxel, then we select the prediction for that voxel with the maximum confidence.
As simple as this strategy might sound, this method has been shown to result in highly robust and efficient outcome in various previous studies \cite{aljabar2009multi, heckemann2006automatic, rohlfing2004evaluation, zhou2017fixed}. The final decision for the $i$-th voxel $y_i^\star$ of ${\hat{\mathbf{Y}}_m}$ is:
\begin{equation}  
\label{equation:ensemble}
y_i^\star = 
    \begin{cases}
        y_i^{V},  & \text{if} \ \exists V, V^\prime \in \{\textup{S},\textup{C},\textup{A}\}, \ V \neq V^\prime \ | \ y_i^{V} = y_i^{V^\prime}  \\
        y_i^{V^\star}, & \text{otherwise}
    \end{cases},
\end{equation}  
where $V^\star = \argmax  \limits_{V \in \{\textup{S},\textup{C},\textup{A}\}} \max\limits_j p^V_{i,j}$.
$p_{i,j}^\text{S}$, $p_{i,j}^\text{C}$, and $p_{i,j}^\text{A}$ denote the probabilities of the $i$-th pixel classified as label $j$ from the sagittal, coronal, and axial planes, respectively.
$y_i^{V}$ denotes the hard estimation for the $i$-th pixel on plane $V$, \emph{i.e.}, $y_i^{V} = \argmax  \limits_{j} p^V_{i,j}$.

As shown in Figure \ref{Fig:fusion}, our multi-planar fusion module improves both over- and under-estimation by fusing aspects from different planes and therefore yields a much better outcome. Note that other rules \cite{asman2013non, wang2018abdominal} can also be easily adapted to this module. We do not focus on discussing the influence of the fusion module in this paper, although intuitively better fusion module should lead to higher performance.

\subsection{Student Model}
\label{Approach:Student}
After generating the pseudo-labels $\{\hat{\mathbf{Y}}_m\}_{m=l+1}^M$ for the unlabeled dataset $\mathcal{S}_{\text{U}}$, the training set can be then enlarged by taking the union of both the labeled and the unlabeled dataset, \emph{i.e.}, $\mathcal{S} = \mathcal{S}_{\text{L}} \cup \{(\mathbf{I}_m, \hat{\mathbf{Y}}_m)\}_{m=l+1}^M$. The student model is trained on this augmented dataset $\mathcal{S}$ the same way we train the teacher model as described in Sec. \ref{baseline}.
The overall training procedure is summarized in Algorithm \ref{Alg:co-training}. In the training stage, we first train a teacher model in a supervised manner and then use it to generate the pseudo-labels for the unlabeled dataset.
Then we alternate the training of the student model and the pseudo-label generation procedures in an iterative manner to optimize the student model $T$ times.
In the testing stage, we follow the method in Sec. \ref{Test} to generate the final estimation using the $T$-th student model.

\section{Experiments}
\label{Experiments}

\begin{algorithm}[t]
\caption{Deep Multi-Planar Co-Training for Multi-Organ Segmentation}
\label{Alg:co-training}
\begin{algorithmic}
   	\Require{A set of labeled data $\mathcal{S}_{\text{L}}=\{(\mathbf{I}_m,\mathbf{Y}_m)\}_{m=1}^l$ and unlabeled volumes $\mathcal{S}_{\text{U}}=\{\mathbf{I}_m\}_{m=l+1}^M$.}
    \Ensure{A trained multi-organ segmentation model $\{\mathbb{M}^\text{S}, \mathbb{M}^\text{C}, \mathbb{M}^\text{A}\}$.}
    \State{$\mathcal{S} \gets{\mathcal{S}_{\text{L}}}$}
    \For{$t = 1$ to $T$}
    	\State{Parse $\mathcal{S}$ into $\mathcal{S}^\text{S}$, $\mathcal{S}^\text{C}$ ,$\mathcal{S}^\text{A}$.}
    	\State{Train $\mathbb{M}^\text{S}$, $\mathbb{M}^\text{C}$, and $\mathbb{M}^\text{A}$ on $\mathcal{S}^\text{S}$, $\mathcal{S}^\text{C}$, and $\mathcal{S}^\text{A}$ respectively.}
        \State{Generate pseudo-class labels $\{\hat{\mathbf{Y}}_m\}_{m=l+1}^M$ for the unlabeled dataset $\mathcal{S}_{\text{U}}$ {by Eq.~\ref{equation:ensemble}}} 
        \State{Augment the training set $\mathcal{S}$ by adding the self-labeled examples to $\mathcal{S}_{\text{L}}$ , \emph{i.e.}, $\mathcal{S} =  \mathcal{S}_{\text{L}} \cup \{(\mathbf{I}_m, \hat{\mathbf{Y}}_m)\}_{m=l+1}^M$.}    
    \EndFor
    \State{Parse $\mathcal{S}$ into $\mathcal{S}^\text{S}$, $\mathcal{S}^\text{C}$ ,$\mathcal{S}^\text{A}$.}
    \State{Train $\mathbb{M}^\text{S}$, $\mathbb{M}^\text{C}$, and $\mathbb{M}^\text{A}$ on $\mathcal{S}^\text{S}$, $\mathcal{S}^\text{C}$, and $\mathcal{S}^\text{A}$ respectively.}
\end{algorithmic}
\end{algorithm}

\subsection{Dataset and Evaluation}
\label{Experiments:DatasetEvalutation}
Our fully-labeled dataset includes 210 contrast-enhanced abdominal clinical
CT images in the portal venous phase, in which we randomly choose 50/30/80 patients for training, validation, and testing, unless otherwise specified. A total of 16 structures (Aorta, Adrenal gland, Celiac AA, Colon, Duodenum, Gallbladder, Interior Vena Cava (IVC), Kidney (left, right), Liver, Pancreas, Superior Mesenteric Artery (SMA), Small bowel, Spleen, Stomach, Veins) for each case were segmented by four experienced radiologists, and confirmed by an independent senior expert.
Our unlabeled dataset consists of 100 unlabeled cases acquired from a local hospital. To the best of our knowledge, this is the largest abdominal CT dataset with the most number of organs segmented.
Each CT volume consists of $319 \sim 1051$ slices of $512 \times 512$ pixels, and have voxel spatial resolution of $([0.523 \sim 0.977] \times [0.523 \sim 0.977]\times 0.5)\textup{mm}^3$.
The metric we use is the Dice-S{\o}rensen Coefficient (DSC), which measures 
the similarity between the prediction voxel set $\mathcal{Z}$ and the ground-truth set $\mathcal{Y}$,
with the mathematical form of ${\mathrm{DSC}\!\left(\mathcal{Z},\mathcal{Y}\right)}=
    {\frac{2\times\left|\mathcal{Z}\cap\mathcal{Y}\right|}{\left|\mathcal{Z}\right|+\left|\mathcal{Y}\right|}}$.
For each organ, we report an average DSC together with the standard deviation over all the testing cases.

\begin{table*}[t!]
\centering
\caption{The comparison of segmentation accuracy (DSC, $\%$) by using $50$ labeled data and varying the number of unlabeled data (\emph{e.g.}, 50-0 indicates $50$ labeled data and $0$ unlabeled data). We report the mean and standard deviation over 80 cases. The \emph{p}-values for testing significant difference between DMPCT (50-100) and FCN (50-0) are shown. Significant statistical improvement is shown in \emph{italic} with $p<0.05$. See Section~\ref{Experiments:Results} for definitions of FCN, SPSL, and DMPCT (Ours).}
\label{Tab:Results}
\begin{tabular}{lcccccl}
\toprule
\multirow{2}{*}{Organ Type} & FCN  & \multicolumn{2}{c}{SPSL} & \multicolumn{2}{c}{DMPCT (Ours)} & \multirow{2}{*}{\emph{p}-value}\\
\cmidrule(lr){2-2} \cmidrule(lr){3-4}  \cmidrule(lr){5-6}
 & 50 - 0 & 50 - 50 & 50 - 100 & 50 - 50 & 50 - 100 & \\
\midrule
Aorta           & $89.14\pm7.95$ & $91.10\pm5.52$ & $90.76\pm5.90$ & $91.43\pm4.88$ & $91.54\pm4.65$ & $\mathit{3.32\times 10^{-5}}$\\
Adrenal gland   & $26.45\pm12.1$ & $29.92\pm14.7$ & $26.93\pm15.6$ & $30.58\pm12.7$ & $35.48\pm11.8$ & $\mathit{1.98\times 10^{-15}}$\\
Celiac AA       & $35.01\pm19.7$ & $37.27\pm19.0$ & $39.78\pm18.4$ & $36.25\pm20.5$ & $40.50\pm18.9$ & $\mathit{1.00\times 10^{-5}}$\\
Colon           & $71.81\pm14.9$ & $78.28\pm13.0$ & $79.58\pm12.9$ & $79.61\pm12.3$ & $80.53\pm11.6$ & $\mathit{7.69\times 10^{-12}}$\\
Duodenum        & $54.89\pm15.5$ & $57.77\pm17.3$ & $62.22\pm14.8$ & $66.95\pm12.6$ & $64.78\pm13.8$ & $\mathit{1.95\times 10^{-19}}$\\
Gallbladder     & $86.53\pm6.21$ & $87.87\pm5.45$ & $88.02\pm5.83$ & $88.45\pm5.07$ & $87.77\pm6.29$ & $\mathit{0.002}$\\
IVC             & $77.67\pm9.49$ & $81.28\pm8.87$ & $82.63\pm7.31$ & $83.49\pm6.94$ & $83.43\pm7.02$ & $\mathit{9.30\times 10^{-14}}$\\
Kidney (L)      & $95.12\pm5.01$ & $95.59\pm4.97$ & $95.88\pm3.68$ & $95.82\pm3.60$ & $96.09\pm3.42$ & $\mathit{3.69\times 10^{-6}}$\\
Kidney (R)      & $95.69\pm2.36$ & $95.77\pm4.93$ & $96.14\pm2.94$ & $96.17\pm2.75$ & $96.26\pm2.29$ & $\mathit{1.74\times 10^{-7}}$\\
Liver           & $95.45\pm2.41$ & $96.06\pm0.99$ & $96.07\pm1.03$ & $96.11\pm0.97$ & $96.15\pm0.92$ & $\mathit{0.005}$\\
Pancreas        & $76.49\pm11.6$ & $80.12\pm7.52$ & $80.93\pm6.84$ & $81.46\pm6.32$ & $82.03\pm6.16$ & $\mathit{2.97\times 10^{-8}}$\\
SMA             & $52.26\pm17.1$ & $51.81\pm18.2$ & $51.94\pm17.1$ & $49.40\pm19.2$ & $52.70\pm17.7$ & $0.667$\\
Small bowel     & $71.13\pm13.1$ & $78.93\pm12.6$ & $79.97\pm12.8$ & $79.49\pm12.1$ & $79.25\pm12.6$ & $\mathit{2.53\times 10^{-22}}$\\
Spleen          & $94.81\pm2.64$ & $95.46\pm2.09$ & $95.58\pm1.90$ & $95.73\pm2.03$ & $95.98\pm1.59$ & $\mathit{1.83\times 10^{-10}}$\\
Stomach         & $91.38\pm3.94$ & $92.62\pm3.71$ & $92.92\pm3.65$ & $93.33\pm3.47$ & $93.42\pm3.21$ & $\mathit{3.30\times 10^{-23}}$\\
Veins           & $64.75\pm15.4$ & $70.43\pm14.3$ & $69.66\pm14.6$ & $69.82\pm14.5$ & $70.23\pm14.4$ & $\mathit{4.16\times 10^{-15}}$\\
\midrule
Mean            & $73.71\pm9.97$ & $76.32\pm9.58$ & $76.87\pm9.08$ & {$77.20\pm8.75$}  & $\mathbf{77.94}\pm8.51$ & $\mathit{4.74\times 10^{-90}}$\\
\bottomrule
\end{tabular}
\end{table*}



\begin{figure*}[t]
\begin{center}
    \includegraphics[width=\linewidth]{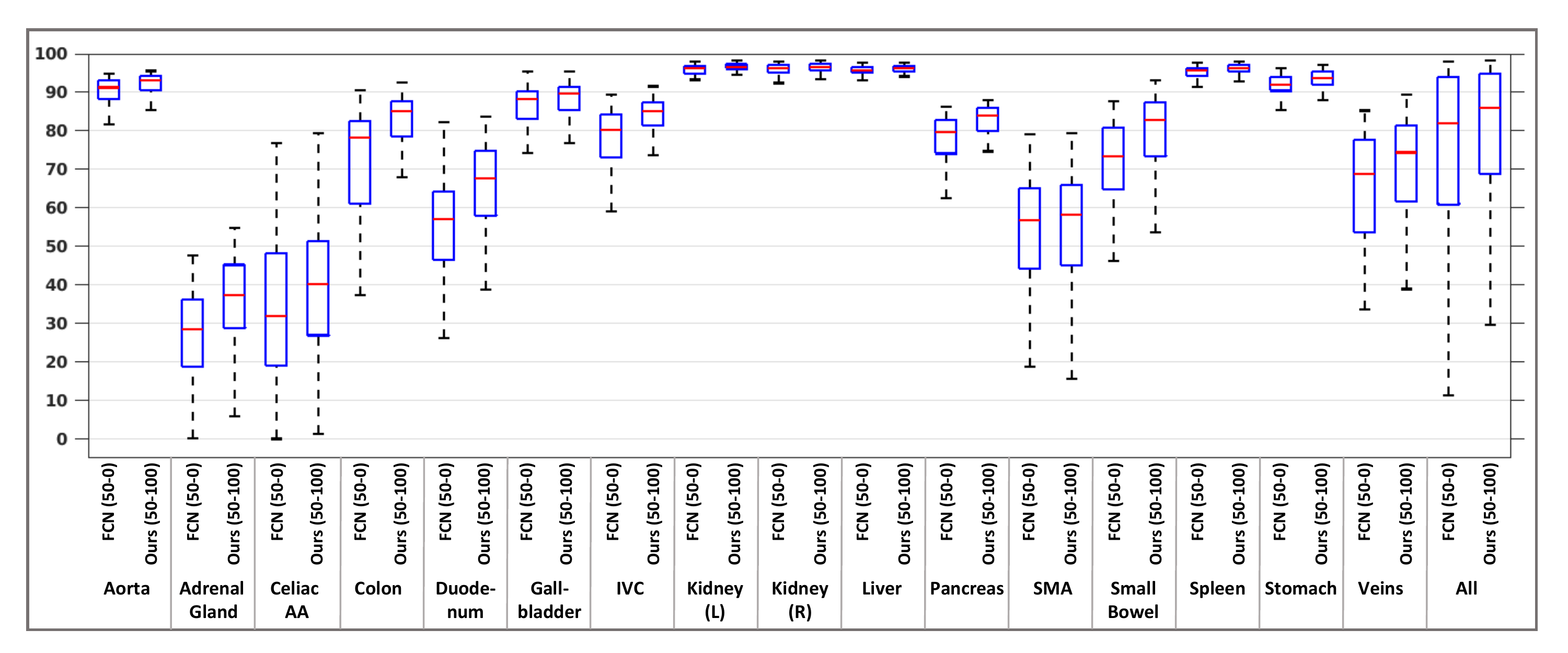}
\end{center}
\vspace{-1ex}
\caption{Performance comparison (DSC, $\%$) in box plots of 16 organs by using 50 labeled data and varying the number of unlabeled data (\emph{e.g.}, 50-0 indicates $50$ labeled data and $0$ unlabeled data). See Section~\ref{Experiments:Results} for definitions of FCN and DMPCT (Ours).}
\label{Fig:boxplot}
\end{figure*}

\begin{figure*}[t]
\begin{center}
    \includegraphics[width=0.82\linewidth]{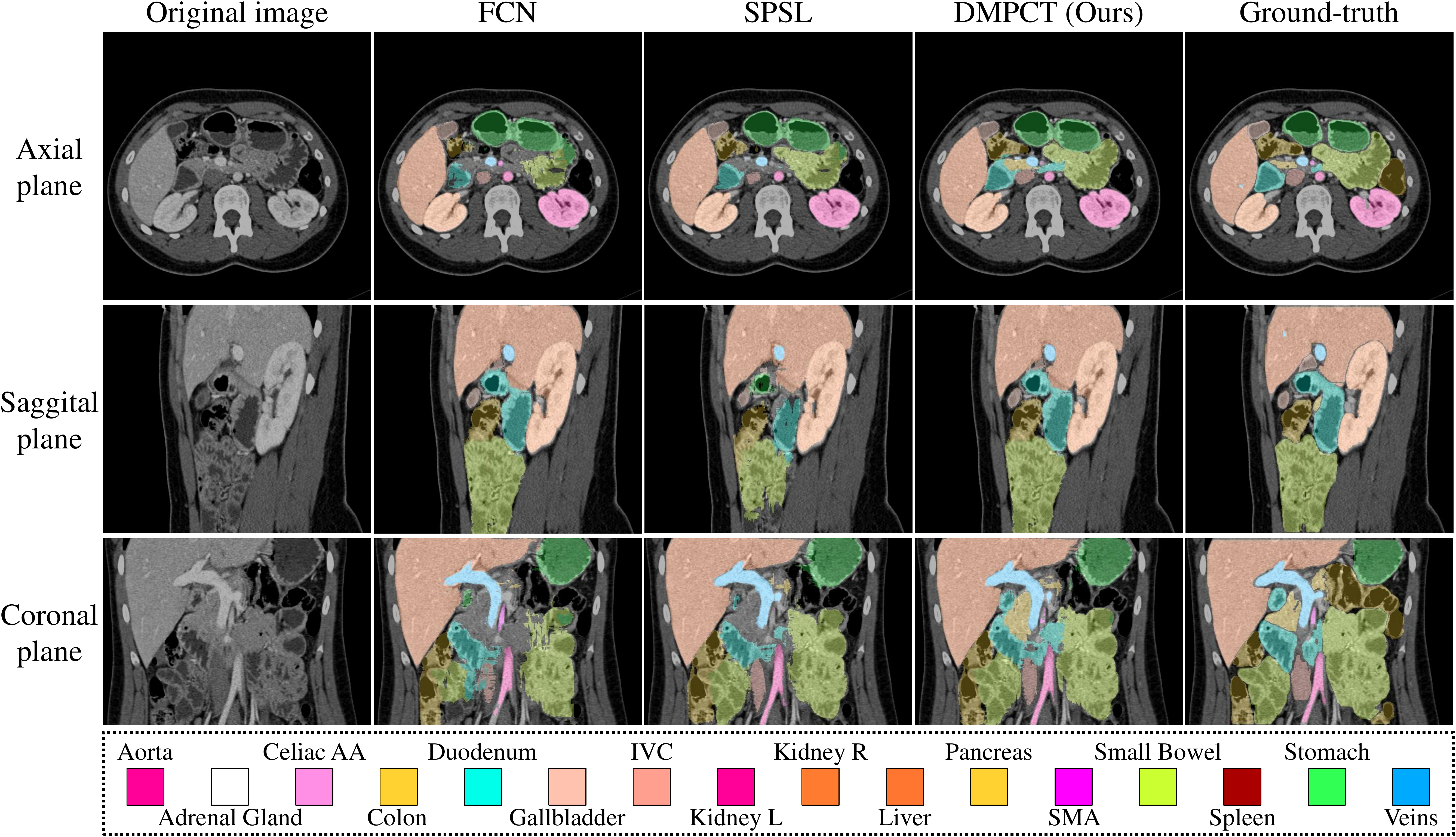}
\end{center}
\vspace{-1ex}
\caption{
	Comparisons among FCN, SPSL, and DMPCT (Ours) viewed from multiple planes.
    50 labeled cases are used for all methods.
    100 unlabeled cases are used for the SPSL and DMPCT.
    For this particular case, FCN obtains an average DSC of $72.75\%$, SPSL gets $78.87\%$, and DMPCT (Ours) gets $80.75\%$. 
    See Section~\ref{Experiments:Results} for definitions of FCN, SPSL, and DMPCT (Ours).
    Best viewed in color.}
\label{Fig:vis}
\end{figure*}

\subsection{Implementation Details}
\label{sec:imp_details}

We set the learning rate to be $10^{-9}$. The teacher model and the student model are trained for $80,000$ and $160,000$ iterations respectively. The validation set is used for tuning the hyper-parameters.
Similar to \cite{harrison2017progressive}, we use three windows of $[-125,\ 275]$, $[-160,\ 240]$, and $[-1000,\ 1000]$ Housefield Units as the three input channels respectively. The intensities of each slice are rescaled to [0.0, 1.0].
Similar to \cite{zhou2017fixed, yurecurrent, wang2018abdominal}, we initialize the network parameters $\mathcal{\theta}$ by using the FCN-8s model \cite{Long_2015_Fully} pre-trained on the PascalVOC image segmentation dataset.
The iteration number $T$ in Algorithm~\ref{Alg:co-training} is set to 2, \emph{i.e.}, $T = 2$, as the performance of the validation set gets saturated.

\subsection{Comparison with the Baseline}
\label{Experiments:Results}

We show that our proposed DMPCT works better than other methods: 1) fully supervised learning method \cite{Long_2015_Fully} (denoted as FCN), and 2) single planar based semi-supervised learning approach \cite{bai2017semi} (denoted as SPSL). Both 1) and 2) are applied on each individual plane separately, and then the final result is obtained via multi-planar fusion (see Sec \ref{Test}). As shown in Table~\ref{Tab:Results}, with $50$ labeled data, by varying the number of unlabeled data from 0 to 100, the average DSC of DMPCT increases from $73.71\%$ to $77.94\%$ and the standard deviation decreases from $9.97\%$ to $8.51\%$. Compared with SPSL, our proposed DMPCT can boost the performance in both settings (\emph{i.e.}, 50 labeled data + 50 unlabeled data and 50 labeled data + 100 unlabeled data). Besides, the \emph{p}-values for testing significant difference between our DMPCT (50 labeled data + 100 unlabeled data) and FCN (50 labeled data + 0 unlabeled data) for $16$ organs are shown in the last column of Table~\ref{Tab:Results}, which suggests significant statistical improvements among almost all organs. Figure \ref{Fig:boxplot} shows comparison results of our DMPCT and the fully supervised method by box plots.

It is noteworthy that greater improvements are observed especially for those difficult organs, \emph{i.e.}, organs either small in sizes or with complex geometric characteristics. Table \ref{Tab:Results} indicates that our DMPCT approach boosts the segmentation performance of these small hard organs by $5.54\%$ (Pancreas), $8.72\%$ (Colon), $9.89\%$ (Duodenum), $8.12\%$ (Small bowels) and $5.48\%$ (Veins), $5.76\%$ (IVC). This promising result indicates that our method distills a reasonable amount of knowledge from the unlabeled data. An example is shown in Figure~\ref{Fig:vis}. In this particular case, the DSCs for Celiac AA, Colon, Duodenum, IVC, Pancreas and Veins are boosted from $60.13\%$, $46.79\%$, $71.08\%$, $69.23\%$, $63.48\%$ to $79.45\%$, $83.81\%$, $77.59\%$, $74.75\%$, $75.31\%$ respectively.


\begin{figure*}[t]
\begin{center}
    \includegraphics[width=0.82\linewidth]{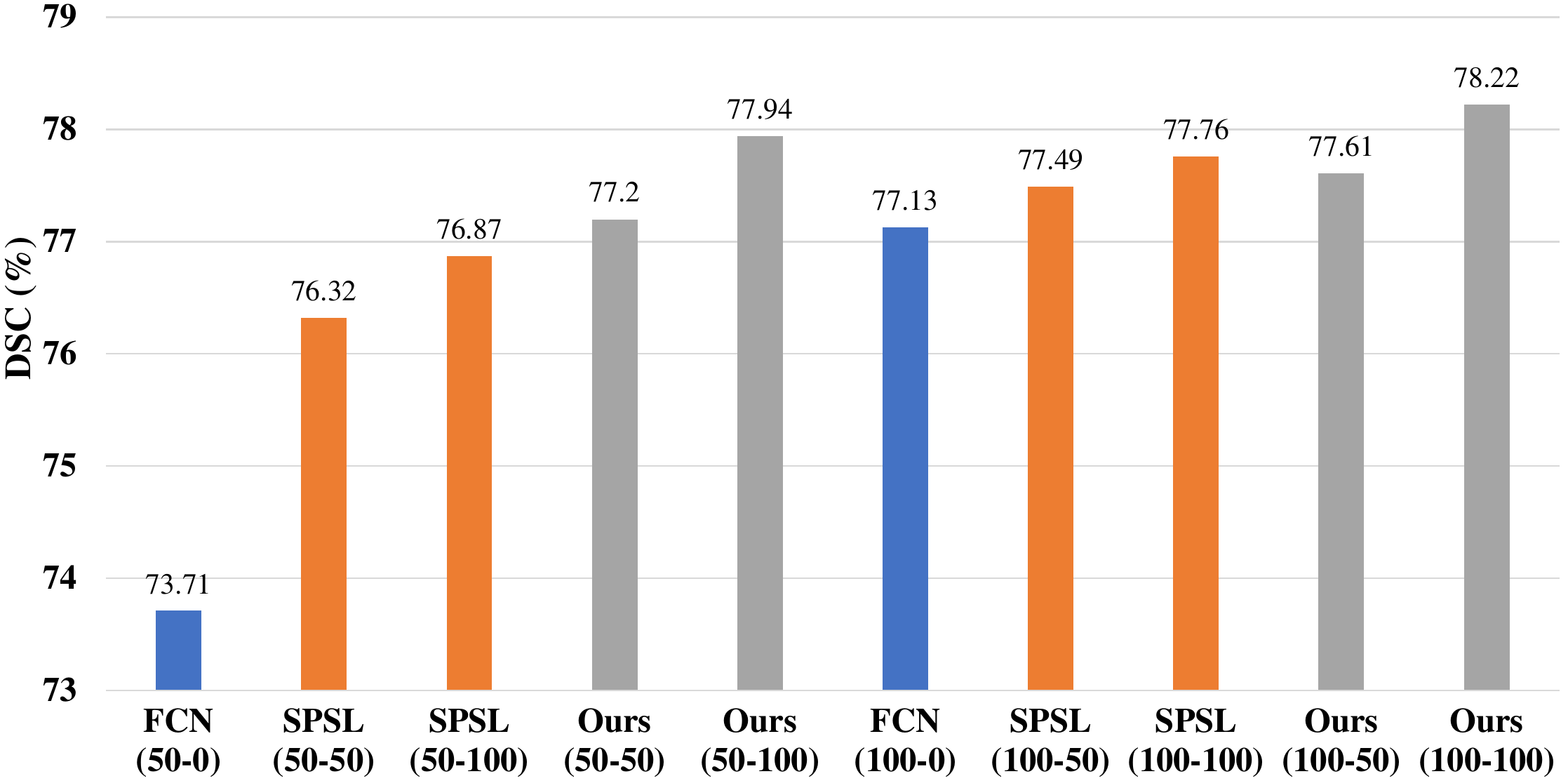}
\end{center}
\vspace{-1ex}
\caption{Ablation study on numbers of labeled data and unlabeled data. Mean DSC of all testing cases under all settings (\emph{e.g.}, 50-0 indicates $50$ labeled data and $0$ unlabeled data). See Section~\ref{Experiments:Results} for definitions of FCN, SPSL, and DMPCT (Ours).}
\label{Fig:ablation}
\end{figure*}

\begin{table*}[t]
\centering
\caption{Cross-dataset generalization results.}
\label{abdomenChallenge}
\begin{tabular}{lccccc}
\toprule
Organ & Spleen      & Kidney (R)   & Kidney (L)   & Gall Bladder   & Liver        \\
\midrule
FCN   & $71.85 \pm 26.13$ & $54.44 \pm 20.04$ & $54.98 \pm 26.63$ & $48.13 \pm 26.07$  & $85.46 \pm 16.81$  \\ 
DMPCT (Ours) & $\mathbf{83.68} \pm 16.53$ & $\mathbf{71.36} \pm 20.85$ & $\mathbf{69.95} \pm 20.50$ & $\mathbf{60.05} \pm 26.91$  & $\mathbf{92.11}\pm 6.46$ \\ \midrule
Organ      & Stomach     & Aorta       & IVC         & Veins       & Pancreas \\
\midrule
FCN & $38.89 \pm 23.86$ & $70.43 \pm 19.70$ & $53.67 \pm 18.40$ & $35.54 \pm 18.94$ & $39.40 \pm 25.34$\\
DMPCT (Ours) & $\mathbf{54.78} \pm 26.57$ & $\mathbf{76.05} \pm 15.99$ & $\mathbf{68.18} \pm 14.58$ & $\mathbf{37.52}\pm 15.86$ & $\mathbf{60.05} \pm 16.61$ \\
\bottomrule
\end{tabular}
\end{table*}

\subsection{Discussion}
\subsubsection{Amount of labeled data}
For ablation analysis, we enlarge the labeled training set to 100 cases and keep the rest of the settings the same. 
As shown in Figure \ref{Fig:ablation}, with more labeled data, the semi-supervised methods (DMPCT, SPSL) still obtain better performance than the supervised method (FCN), while the performance gain becomes less prominent. This is probably because the network is already trained well when large training set is available. 
We believe that if much more unlabeled data can be provided the performance should go up considerably. 
In addition, we find that DMPCT outperforms SPSL in every setting, which further demonstrates the usefulness of multi-planar fusion in our co-training framework.

\subsubsection{Comparison with 3D network-based self-training}
Various previous studies \cite{Prasoon_2013_Deep, wang2018abdominal} demonstrate that 2D multi-planar fusion outperforms directly 3D learning in the fully supervised setting. 3D CNNs come with an increased number of parameters, significant memory and computational requirements. Due to GPU memory restrictions, these 3D CNN approaches which adopt the sliding-window strategy do not act on the entire 3D CT volume, but instead on local 3D patches \cite{cciccek20163d, dou20163d, gibson2017towards}. This results in the lack of holistic information and low efficiency.  
In order to prove that DMPCT outperforms direct 3D learning in the semi-supervised setting, we also implement a patch-based 3D UNet \cite{cciccek20163d}. 3D UNet gets  $69.66\%$ in terms of mean DSC using 50 labeled data. When adding 100 unlabeled data the performance even drops to $65.21\%$. This clearly shows that in 3D learning the teacher model is not trained well, thus the errors of the pseudo-labels are reinforced during student model training.

\subsubsection{Comparison with traditional co-training}
In order to show that our DMPCT outperforms traditional co-training algorithm \cite{blum1998combining}, we also select only the most confident samples  during each iteration. Here the confidence score is measured by the entropy of probability distribution for each voxel in one slice. Under the setting of 50 labeled cases and 50 unlabeled cases, we select top 5000 samples with the highest confidence in each iteration. The whole training process takes about 6-7 iterations for each plane. The complete training requires more than 50 hours. Compared with our approach, this method requires much more time to converge.
It obtains a mean DSC of $76.52\%$, slightly better than SPSL but worse than our DMPCT, which shows that selecting the most confident samples during training may not be a wise choice for deep network based semi-supervised learning due to its low efficiency.

\subsubsection{Cross dataset generalization}
We apply our trained DMPCT model (50 labeled data + 100 unlabeled data) and baseline FCN model (50 labeled data + 0 unlabeled data) on a public available abdominal CT datasets\footnote{30 training data sets at \url{https://www.synapse.org/\#!Synapse:syn3193805/wiki/217789}} with 13 anatomical structures labeled \emph{without any further re-training} on new data cases. 10 out of the 13 structures are evaluated which are also manually annotated in our own dataset and we find that our proposed method improves the overall mean DSC and also reduces the standard deviation significantly, as shown in Table \ref{abdomenChallenge}. 
The overall mean DSC as well as the standard deviation for the 10 organs is improved from  $59.23 \pm 22.20\%$ to $67.38 \pm 19.64\%$. We also directly test our models on the NIH pancreas segmentation dataset of 82 cases\footnote{\url{https://wiki.cancerimagingarchive.net/display/Public/Pancreas-CT}} and observe that our DMPCT model achieves an average DSC of $66.16\%$, outperforming the fully supervised method, with an average DSC of $58.73\%$, by more than $7\%$. 
This may demonstrate that our approach, which leverages more unlabeled data from multiple planes, turns out to be much more generalizable than the baseline model.

\subsubsection{Computation time}
In our experiments, the teacher model training process takes about $4.94$ hours on an NVIDIA TITAN Xp GPU card for $80,000$ iterations over all the training cases. The average computation time for generating pseudo-label as well as testing per volume depends on the volume of the target structure, and the average computation time for $16$ organs is approximately $4.5$ minutes, which is comparable to other recent methods \cite{zhou2017fixed, Roth_2016_Spatial} even for single structure inference. The student model training process takes about $9.88$ hours for $160,000$ iterations.

\section{Conclusion}
\label{Conclusions}
In this paper, we designed a systematic framework DMPCT for multi-organ segmentation in abdominal
CT scans, which is motivated by the traditional co-training strategy to incorporate multi-planar information for the unlabeled data during training. The pseudo-labels are iteratively updated by inferencing comprehensively on multiple configurations of unlabeled data with a multi-planar fusion module.
We evaluate our approach on our own large newly collected high-quality dataset.
The results show that 1) our method outperforms the fully supervised learning approach by a large margin;
2) it outperforms the single planar method, which further demonstrates the benefit of multi-planar fusion;
3) it can learn better if more unlabeled data provided especially when the scale of labeled data is small. 

Our framework can be practical in assisting radiologists for clinical applications since the annotation of multiple organs in 3D volumes requires massive labor from radiologists. 
Our framework is not specific to a certain structure, but shows robust results in multiple complex anatomical structures within efficient computational time.
It can be anticipated that our algorithm may achieve even higher accuracy if a more powerful backbone network or an advanced fusion algorithm is employed, which we leave as the future work.

\vspace{1ex}\noindent\textbf{Acknowledgement.}~This work was supported by the Lustgarten Foundation for Pancreatic Cancer Research and also supported by NSFC No. 61672336. We thank Prof. Seyoun Park, Dr. Lingxi Xie, Cihang Xie, Zhishuai Zhang, Fengze Liu, Zhuotun Zhu and Yingda Xia for instructive discussions.

{\small
\bibliographystyle{ieee}
\bibliography{egbib}
}

\end{document}